# Semi-automated analysis of audio-recorded lessons: The case of teachers' engaging messages

**Samuel Falcon:** [1]University of Las Palmas de Gran Canaria, Department of Education, [2]Instituto Universitario de Análisis y Aplicaciones Textuales (IATEXT), University of Las Palmas de Gran Canaria. samuel.falcon@ulpgc.es, https://orcid.org/0000-0003-3314-1945

**Carmen Álvarez-Álvarez[3]:** University of Cantabria, Department of Education, carmen.alvarez@unican.es, https://orcid.org/0000-0002-8160-2286

**Jaime Leon\*:** [1]University of Las Palmas de Gran Canaria, Department of Education, [2]Instituto Universitario de Análisis y Aplicaciones Textuales (IATEXT), University of Las Palmas de Gran Canaria. jaime.leon@ulpgc.es, https://orcid.org/0000-0002-9587-4047

**Abstract**

Engaging messages delivered by teachers are a key aspect of the classroom discourse that influences student outcomes. However, improving this communication is challenging due to difficulties in obtaining observations. This study presents a methodology for efficiently extracting actual observations of engaging messages from audio-recorded lessons. We collected 2,477 audio-recorded lessons from 75 teachers over two academic years. Using automatic transcription and keyword-based filtering analysis, we identified and classified engaging messages. This method reduced the information to be analysed by 90%, optimising the time and resources required compared to traditional manual coding. Subsequent descriptive analysis revealed that the most used messages emphasised the future benefits of participating in school activities. In addition, the use of engaging messages decreased as the academic year progressed. This study offers insights for researchers seeking to extract information from teachers' discourse in naturalistic settings and provides useful information for designing interventions to improve teachers' communication strategies.

# 1. Introduction

Teachers' discourse has the power to shape students' outcomes (Caldarella et al., 2023; Howe & Abedin, 2013; Mercer, 2010). In the classroom, teachers' discourse can be broadly divided into instructional time, where the focus is on delivering academic content, and non-instructional time, which involves interactions with students that are not directly related to teaching the curriculum (Dale et al., 2022; Nurmi, 2012; Parsons et al., 2018). Although instructional time is crucial, the interactions during non-instructional time, which constitute about 20% of teachers' discourse (OECD, 2019), also play a significant role in the overall educational experience, impacting various aspects of students. For instance, recent research such as Putwain et al. (2021) found that messages highlighting the consequences of failure prior to high-stakes exams in predicted lower self-determined motivation in secondary education students, which correlated with poorer performance. Conversely, studies of primary school students by Spilt et al. (2016) and Caldarella et al. (2020) found that positive teacher messages were associated with improved student behaviour and classroom management.

In this context, engaging messages have attracted the attention of researchers as they have been linked to students' motivation to learn, performance and well-being (Santana-Monagas et al., 2024; Santana-Monagas, Putwain, et al., 2022; Santana-Monagas & Núñez, 2022). The results of these studies suggest that students whose teachers use more messages highlighting the future benefits or personal satisfaction of engaging in school tasks tend to perform better. However, these studies also found that many teachers often use messages that emphasise the disadvantages of not engaging in the tasks, focusing on punishment, or sometimes do not attempt to engage students with any kind of message at all, leading to poorer results. From this perspective, enhancing the use of the first type of messages could be of great interest, as it can lead to significant improvements in student outcomes (Gregory et al., 2017; Hopfenbeck, 2024; Santana-Monagas & Núñez, 2022).

There are different approaches that aim to improve teachers' discourse (Hill & Grossman, 2013; Schelling & Rubenstein, 2023; Sedova et al., 2016), with feedback based on actual classroom practices being one of the most effective methods (Kucan, 2007; Stein & Matsumura, 2009). However, collecting observational data to provide feedback is a time-consuming task (Rahman, 2016). For example, in studies such as Caldarella et al. (2020), researchers have to spend a considerable amount of time in the classroom observing teachers' behaviour and cannot carry out observations in different classrooms simultaneously. As an alternative, other researchers suggest recording lessons for subsequent observation and extraction of talk feature of interest (Floress et al., 2018). This method allows for thorough assessment and provides the flexibility to revisit and reflect on the data, leading to more informed coding (Vrikki et al., 2019), but reviewing the recording of an entire lesson can be as time-consuming as observing the lesson in situ. For this reason, other authors advocate transcribing lessons into text, a methodology known as transcript-based lesson analysis (Arani, 2017). This methodology can save some time, as reading the information is quicker than listening to it (Rahayu et al., 2020). The efficiency and depth of analysis provided by transcript-based lesson analysis has led to it becoming one of the most widely used methods by researchers to analyse teachers' discourse in recent years (Lefstein et al., 2020; Mullet, 2018).

Despite its advantages, transcript-based lesson analysis also has limitations. For instance, Dale et al. (2022) reported that transcribing a single one-hour lesson could take up to four hours. Consequently, they propose a semi-automated methodology using automatic transcription for analysing teacher discourse in lesson recordings. In their study, they compared the effectiveness of this semi-automated methodology against traditional transcription methods and demonstrated its capability to efficiently and accurately identify key features of teachers' discourse. This approach significantly reduced the time and effort required for transcription while maintaining reliability, thereby offering a scalable solution for large-scale analysis of discourse.

While promising, this semi-automated methodology still requires manual coding, a process that consumes significant time and resources (Boden et al., 2020). In this context, filtering the lesson transcripts with lexical rules could be a significant advance (Ghauth & Abdullah, 2010). This process involves reducing the original transcript size based on keywords, resulting in a final set of sentences that contain those keywords, thereby making the analysis more efficient and manageable (Winarti et al., 2021). Using this method, we would be able to discard the parts of the lesson devoted to instruction and code only the teacher's discourse dedicated to interaction with students, where engaging messages are concentrated (Falcon et al., 2023; Santana-Monagas & Núñez, 2022). In this way, we would be able to save large amounts of time in manually coding transcripts and analyse larger amounts of data.

This study builds on the semi-automated discourse analysis proposed by Dale et al. (2022) and seeks to enhance it by optimising manual coding using keyword-based filtering within the framework of teachers' engaging messages. Therefore, the study has a dual objective: (1) to create a method that, in addition to automatic transcription, includes a filtering process based on keywords to facilitate coding, and (2) to use this method to extract observations of the engaging messages used by teachers. These objectives aim to enable the collection of large-scale real observations of engaging messages, allowing for the study of the most common messages, how their use varies across educational levels, and how their use changes throughout the academic year. Drawing from these objectives, this study proposes the following research questions:

RQ1: How effective is the proposed method in streamlining the coding process of engaging messages compared to traditional methods?

RQ2: What types of engaging messages are most frequently used by teachers?

RQ3: How does the use of engaging messages by teachers change across different educational levels and over the course of an academic year?

The results will determine the efficacy of this method in obtaining observations of engaging messages more efficiently, facilitating the analysis of these messages from large datasets. Additionally, if the methodology proves effective, it could be used to study other types of messages or aspects of teacher discourse more easily. Furthermore, the direct observations will provide insights into the actual use and variability of engaging messages, potentially paving the way for interventions to refine teachers' communication approaches (Cañadas, 2021). Given that teacher training has positively influenced student attitudes and behaviours (Allen et al., 2011; Caldarella et al., 2015; Gregory et al., 2017), such interventions might further enhance students' engagement and performance.

### *1.1 Teachers' engaging messages*

Teachers' engaging messages are those that teachers use to engage their students in school tasks (Falcon & Leon, 2023; Santana-Monagas, Putwain, et al., 2022; Santana-Monagas & Núñez, 2022). These messages are characterized by focusing on the positive consequences of engagement or by emphasizing the disadvantages of not engaging. In addition, teachers can appeal to different types of motivational incentives while attempting to engage their students. Therefore, engaging messages fall under the umbrella of two theories: Message Framing Theory (MFT; Rothman & Salovey, 1997) and Self-Determination Theory (SDT; Deci & Ryan, 2016; Ryan & Deci, 2020).

According to the MFT, messages can generate different outcomes depending on whether they highlight the benefits of doing a specific activity (*gain-framed*) or highlight the unfavourable consequences of not engaging in the activity (loss-framed; Rothman & Salovey, 1997). Drawing on this theory, research suggests that *gain-framed* messages are more effective in engaging individuals (Tversky & Kahneman, 1986). For instance, in the medical context, findings show that if the doctor emphasises the benefits of having surgery, the patient is more likely to undergo the procedure than if he emphasises the possible disadvantages (Rothman et al., 2006). In education, there is evidence that *loss-framed* messages, more common near high-stakes exams, can diminish student motivation and performance (Putwain et al., 2021; Putwain & Best, 2011; Putwain & Symes, 2011). From this perspective, it would be useful for teachers to use *gain-framed* messages to engage learners in school tasks instead of *loss-framed* messages.

These messages can also appeal to different types of motivational incentives to engage students (Falcon & Leon, 2023; Santana-Monagas & Núñez, 2022). As stated by the SDT (Deci & Ryan, 2016; Ryan & Deci, 2020), there is a continuum on which we can identify different types of motivation, ranging from externally to internally sourced. Externally driven motivations might involve rewards or punishments (*extrinsic*) or relate to personal or others' emotions (*introjected*). Conversely, internal motivations see individuals valuing the task (*identified*) or deriving pleasure from it (*intrinsic*). Evidence suggests students excel and are more engaged when intrinsically motivated (Ryan & Patrick, 2001; Taylor et al., 2014). For this reason, it would be beneficial if teachers used messages that appeal to *identified* or *intrinsic* motivations, as that will better engage students (Moreno et al., 2021).

MFT and SDT, when integrated, offer a more comprehensive perspective on this subject. MFT overlooks the specific types of motivation within a message that might influence learners' outcomes. Conversely, SDT doesn't consider the framing of a message when teachers use a particular motivational approach, even if this framing affects student results. For further theoretical consolidation, Gigerenzer (2017) proposes to work by integrating different theories. Following his recommendations, prior studies integrated both theories to improve the analysis of teachers' messages (Falcon & Leon, 2023; Santana-Monagas, Núñez, et al., 2022; Santana-Monagas & Núñez, 2022), resulting in the development of teachers' engaging messages and their eight categories (Table 1).

[Insert Table 1]

Previous studies on teachers' engaging messages have demonstrated that *gain-framed*, *identified* and *intrinsic* messages positively predict secondary student motivation to learn and performance (Santana-Monagas, Putwain, et al., 2022). Additionally, this type of message has been found to positively influence teacher-student relatedness student vitality and well-being (Santana-Monagas et al., 2023; Santana-Monagas & Núñez, 2022). Further research has identified associations between teachers' perceived autonomy and their use of engaging messages, revealing distinct patterns in message (Santana-Monagas, Núñez, et al., 2022). These findings indicated that students whose teachers favoured *gain-framed* messages exhibited superior academic performance.

These previous results highlight the potential benefits of developing interventions aimed at enhancing teachers' use of engaging messages, as they could positively impact various aspects of the educational experience (Allen et al., 2011; Gregory et al., 2017). Among all types of interventions, those based on providing feedback on teachers' actual use of messages appear to be the most promising for achieving real improvements (Kucan, 2007; Laranjo et al., 2021; Stein & Matsumura, 2009). Therefore, in order to obtain observations of engaging messages more efficiently to provide feedback more easily, this study aims to extend the semi-automated methodology proposed by Dale et al. (2022) for analysing teacher discourse by including a keyword-based filtering step (Winarti et al., 2021). In addition, a descriptive analysis of the observed messages will be carried out to gain insights into their general use and how they vary according to the level of education and during the academic year.

## 2. Method

### *2.1 Participants*

Over two academic years, 75 teachers (42 females, 33 males; mean age=45.01, SD=8.69) from public secondary schools in five regions of Spain participated in the study. Teachers could choose to participate with one or more of their groups, resulting in a total of 126 participating groups. The study covered different educational levels, ranging from grades 9 to 12. In Spain, secondary education, which covers grades 7-10, is compulsory and offers a general curriculum with the possibility of electives in science or humanities. Upper secondary education, grades 11-12, is not compulsory and prepares students for university, with the possibility of specialisation in science or humanities. The distribution of participating groups by level of education was as follows: grade 9 (41), grade 10 (30), grade 11 (18) and grade 12 (37).

### *2.2 Procedure*

At the beginning of each academic year, we contacted school principals to recruit participants. We clarified the study's objectives, assured participants of their voluntary and confidential involvement, and obtained their consent through an informed consent form. Our research adhered to both national and European data protection standards.

Over the course of two academic years, we asked teachers to record at least eight lessons per trimester using small (8 cm) recorders provided by us. These recorders were set to capture only nearby voices, ensuring that only the teacher's voice was recorded. We trained teachers on how to use the recorders and requested audio submissions at the end of each trimester. The final number of audio recordings received was 2,477. The difference in the number of recordings expected and received was due to various reasons, including medical leaves during one or more terms and the forgetting to record some lessons.

Upon receiving the audio recordings, we automatically transcribed them using a cloud-based service. Following the guidelines of Dale et al. (2022), we configured the transcription segmentation so that each segment was separated by a pause of at least one second. This minimised the number of incomplete or broken sentences and helped us in. After transcribing the audios, we applied the keyword-based filtering to the transcripts, and we coded the filtered transcripts to identify and categorise the engaging messages.

## *2.3 The coding process of engaging messages*

The process followed to improve manual codification of engaging messages integrating the keyword-based filtering involved three steps (Figure 1) and two python programs[1].

---

[1] https://osf.io/ms7n3/?view_only=637da679c2c34ff68bd83683f4397fc5.

[Insert Figure 1]

*Step 1 - Initial coding*: In the first trimester of the first academic year, we obtained 61 audios (approximately 750 pages of text after the transcription). In this first step, we performed a traditional manual codification reading all the text pages. To identify the engaging messages in the transcripts, we trained two research assistants in the use of a codebook with theory-based examples of all types of messages (similar to Table 1). Following Strijbos et al. (2006) guidelines for coding information in a content analysis, we included additional instructions in the codebook such as selecting messages from the teacher that: (1) were aimed at engaging students in school tasks, (2) had a frame, either *gain* or *loss*, (3) appealed to a motivational incentive, and (4) were meaningful in their own sense (could be one or more sentences). The manual coding in this first step allowed to identify 168 engaging messages.

*Step 2 – Development of the keyword list and transcript filtering*: Following the methodology used by Winarti et al. (2021) for keyword filtering, we created a python program to compare all the words in the 168 engaging messages identified in the previous step with those in the rest of the transcripts. This comparison allowed us to classify words according to whether they were frequent in the engaging messages but rare in other parts of the transcripts, identifying them as keywords. Some of these keywords included: "*study*", "*exam*", "*work*", "*future*", "*qualifications*", "*stress*", "*proud*", "*money*", and "*positive*". We then established multiple keyword lists (100-150 words each) and conducted several keyword-based filtering on the previously analysed transcripts.

To determine the optimal keyword list, assistants examined these filtered transcripts. We selected a final list of 111 keywords because it allowed us to obtain filtered transcripts that contained 100% of the messages manually identified in the first step. Moreover, this list enabled us to reduce the number of pages to analyze to just 75, effectively discarding 90% of the original transcripts. This reduction significantly improved message identification efficiency and minimized the risk of oversight, resulting in a total of 178 engaging messages identified in the first trimester of the first academic year. The reliability of message identification showed a satisfactory inter-coder agreement of 98.71% (O'Connor & Joffe, 2020).

*Final coding*: The last step comprised the identification of engaging messages in the rest of the transcripts. For this, we first applied the filtered the remaining transcripts with the 111-keyword list, reducing their length. After that, the assistants used the codebook to classify the messages in the different categories. Reliability results showed very good (98.18% for the category "*intrinsic*" of the *appeal* dimension) to acceptable (74.40% for the category *"identified"* of the *appeal* dimension) agreements (Hartmann, 1977; Stemler, 2004) with most categories above 80%.

## *2.4 Data analysis*

We conducted three descriptive data analyses: overall results, results across educational levels and results throughout the trimesters. For the first and third cases, we calculated the percentage of messages used from each category based on the total message count. However, as the number of groups at each educational level was different, we first calculated the following ratio for each type of messages: *message count / number of groups*. Then, we calculated the percentages of each category of messages based on that ratio.

## 3. Results

### *3.1 Overall results*

After examining the recordings from two academic years, we identified 856 messages. These were classified into eight categories based on frame (*gain-* or *loss-framed*) and appeal (*extrinsic*, *introjected*, *identified*, and *intrinsic*) dimensions. In Table 2, we show examples of real engaging messages obtained after codifying the audio files:

[Insert Table 2]

For the overall analysis, we examined the percentage of each of the message categories based on the total number of messages detected over the two academic years (Figure 2).

[Insert Figure 2]

The first thing to highlight from this figure is that there were not many *intrinsic* messages, either *gain-* or *loss-framed.* However, we can observe that teachers rely most frequently on *identified* messages, both *gain-* and *loss-framed*. After these, the engaging messages most frequently used by teachers are *loss-framed extrinsic*, *gain-framed extrinsic*, *loss-framed introjected*, and *gain-framed introjected*, in this order.

Furthermore, obtaining real observations of engaging messages allowed a deeper examination of the motivational incentives used. For external motivation sources, regardless of the frame, *extrinsic* messages highlighted three aspects: free time, study duration, and tangible rewards/punishments. Teachers guided students to study to enjoy more leisure or warned of increased study and reduced hobbies if not engaged. Similarly, students were prompted to study consistently to avoid cramming. Lastly, teachers leveraged various incentives, from attitude points and treats to meetings with school management. Conversely, *introjected* messages focused on the emotions students might feel based on their engagement, influenced by their feelings, those of the teacher, or family members.

Concerning internal sources of motivation, *identified* messages resonated with motives relatable to learners, leading to varied content based on the student group and whether the teacher addressed the class or an individual. Predominantly, these messages centred around themes of learning, prospective studies, and the broader future. Lastly, *intrinsic* messages were not very frequent, but mainly appealed to activities done for "the student's own sake", based on interest and enjoyment or loss of that pleasurable experience.

### *3.2 Results across educational levels*

The total number of messages across educational levels were as follows: grade 9 = 302, grade 10 = 176, grade 11 = 157, and grade 12 = 221. However, as there were not the same number of groups in each level, we first calculated the ratio and then the percentage, as explained in Section 2.4. The differences between levels are displayed in the figure below (Figure 3).

[Insert Figure 3]

We can see how most *extrinsic* and *introjected* messages (both *gain-* and *loss-framed*) are mostly concentrated in grade 9 and 11. *Identified* messages, on their side, are mostly concentrated in grade 11, but then appear to be more distributed across the rest of the grades. What is interesting about the data in this figure is that *identified* messages (both *gain-* and *loss-framed*) were the most common messages in all educational levels. In addition, many of the *loss-framed identified* messages detected in grade 12 referred to future studies. It is important to note that, at the end of this grade, students have to take a university entrance exam and the grade obtained will determine to a large extent the university options they will have access to. Finally, almost all the *intrinsic* messages were *gain-framed* and were evenly distributed across all grades.

### *3.3 Results throughout the academic year*

The total number of messages throughout the trimesters were as follows: first trimester = 353, second trimester = 331, and third trimester = 172. Differences between trimesters are displayed below (Figure 4).

[Insert Figure 4]

The general trend in the data seems to be a decrease in the use of engaging messages, in almost all categories, as the academic year progresses. This trend is particularly clear in the case of *gain-framed identified* messages but is less strong for other categories. For instance, *gain-* and *loss-framed extrinsic* and *introjected*, and *loss-framed identified* messages experienced an increase in the number of messages detected in the second trimester, only to fall in the third trimester to lower levels than in the first trimester. Interestingly, it seems that *identified* messages are also the most used throughout all the trimesters.

## 4. Discussion

In this study, we investigated a new semi-automatic method derived from the combination of the approach proposed by Dale et al. (2022) with keyword filtering for analysing secondary teachers' use of engaging messages within their discourse. The aims were to investigate how effective was the proposed method in streamlining the coding process compared to traditional methods, to observe what types of engaging messages were most frequently used by teachers, and to examine how did the use of engaging messages by teachers change across different educational levels and over the course of the academic year.

*4.1 Semi-automatic methodology*

In response to RQ1, the methodology employed to collect direct observations of engaging messages yielded satisfactory results. It enabled us to analyse a substantial volume of data, 2,477 audio-recorded lessons over two academic years, within a short timeframe, a feat unachievable with traditional coding methods. This success was primarily due to the keyword-filtering step, which allowed us to review only 10% of the total information. This approach not only saved time but also proved more effective. A comparison of traditional coding and coding after the keyword-filtering on data from the first trimester of the first year showed that assistants identified more messages using the new methodology. As discussed, this improvement is likely due to reduced decision fatigue (Vohs et al., 2005), as reading 750 pages of text to find engaging messages is significantly more demanding than reading only 75.

Our contribution in the methodological aspect is twofold. Firstly, we have refined a recent method (Dale et al., 2022) that optimises teacher discourse analysis by integrating keyword-filtering, a proven effective method for information filtering (Winarti et al., 2021). Secondly, we have documented every step and decision made during this development, enabling other researchers to replicate our approach. This paves the way for more optimised studies of various aspects of teacher discourse from audio recordings, both instructional, such as those examined by Dale et al. (2022), and non-instructional, such as fear appeals (Putwain et al., 2017), motivational messages (Ahmadi et al., 2023; Putwain et al., 2021), praise (Jenkins et al., 2015; Lipnevich et al., 2023) and reprimands (Caldarella et al., 2023), among many others .

*4.2 Prevalence of engaging messages*

Regarding RQ2, we were able to observe the prevalent use of *gain-framed identified* messages. Given the MFT, SDT, and prior research on engaging messages, this is promising, as messages that emphasises the benefits of engaging in school tasks appealing to internal motivation tend to enhance student engagement and outcomes (Moreno et al., 2021; Ryan & Deci, 2020) These results are promising as they suggest that teachers are using adequate strategies to foster better engagement among students.

While *gain-framed identified* messages were the most common, it is important to note the significant presence of *loss-framed identified* messages as the second most common category. While generally associated with increased anxiety and subpar results (Jang & Feng, 2018; Putwain & Remedios, 2014), recent studies suggest that if students perceive such messages as challenges, they can be engaging (Putwain et al., 2023). Moreover, while *gain-framed* messages excel in low-risk scenarios, *loss-framed* ones might be more effective for higher-risk situations (Jang & Feng, 2018). These results are drawn from studies that use students' perceptions of teachers' use of messages as a method of collection. Therefore, further research should be conducted combining this new semi-automatic method of data. This approach could validate and enhance our understanding of how engaging messages influences student motivation and performance.

### *4.3 Differences in educational levels and across the academic year*

Answering RQ3, the educational level where we observed the highest prevalence of engaging messages was grade 11. Recognised as a crucial transitional year between mandatory and optional secondary education, teachers in this grade may be especially driven to optimise student engagement (Jimerson & Haddock, 2015). Their intent might stem from a desire to adequately prepare students for subsequent educational stages. Another interesting level to analyse is grade 9, which contained the youngest pupils in the study. Despite the predominance of *identified* messages, teachers at this level also engage their students by using a large number of *extrinsic* messages, especially *loss-*

*framed* ones. They rely more on rewards and punishments to engage their students in school tasks. Appealing to external forms of motivation is a first step in motivating students and facilitating the internalisation of these motivations into more self-determined ones (Ryan & Deci, 2020). However, teachers should be careful with such messages, as their overuse can lead to a weakening effect on the motivation to engage in tasks (Deci, 1971).

Regarding the temporal distribution across the academic year, there is a clear trend towards a decrease in engaging messages in general. A plausible explanation for this might be that the time teachers spend on non-instructional discourse decreases towards the end of the academic year, as teachers focus more on completing the curriculum, leading to a decrease in the use of engaging messages. The decline is especially marked for *gain-framed identified* messages, with other messages showing a less notable decrease.

The insights gained from these observations contribute to a better understanding of teachers' use of engaging messages and can inform future studies aimed at improving their use. Specifically, these findings argue for interventions that promote the greater use of *intrinsic* messages, which have been the least used even though they could be the most beneficial for students (Ryan & Deci, 2020). Additionally, tailored interventions can be designed for different educational levels, such as encouraging more *identified* messages in grade 9 and more *gain-framed* messages in grade 12, while also ensuring that these messages are consistently used throughout the academic year.

### *4.4 Limitations and future perspectives*

Despite our contributions, this study has some limitations. Firstly, although the coding process was improved, it still required the assistance of two coding assistants and remained time-consuming. Consequently, we requested only eight lessons per group from each teacher, which limited the scope of our results. Although we examined a large number of audio recordings, a more efficient coding method could have significantly increased the data size. In this context, it is necessary to mention the recent advancements in large language models, which have emerged as promising options for automatic text analysis (Demszky et al., 2023). Future studies could take the methodology presented in this work and compare it with methodologies employing these models to determine if they allow for a more optimal analysis of teacher discourse.

While these findings are pertinent to Spanish teachers, it is crucial to approach them with caution. Cultural variances in teacher motivation and engagement strategies have been noted in prior research (Hagger et al., 2007). To discern potential differences in the use of engaging messages, a broader cross-cultural study encompassing teachers from varied nations is essential.

The study did not include any measure of teachers' intentions and students' perceptions. In this study we have limited ourselves to developing the method for identifying and coding the engaging messages and analysing their content and evolution. However, future studies could examine the relations between these observational engaging messages, teachers' intentions with these messages, and students' perceptions of them.

This study's findings offer valuable insights for future work on devising interventions that encourage engaging message use among secondary school teachers. Interventions targeting teachers have been shown to yield positive student outcomes (Allen et al., 2011; Gregory et al., 2017). Additionally, the refined coding procedure could facilitate rapid feedback for teachers regarding their actual message use. Such feedback has the potential to be an effective intervention method, given its proven capacity to alter behaviour (Kucan, 2007; Sedova et al., 2016; Stein & Matsumura, 2009). Therefore, further research should be carried out to establish the effectiveness of an intervention that aims to educate teachers on which messages to use to engage students in school tasks and analyse it effects on students outcomes.

**5. Conclusions**

In this study we developed an optimised semi-automatic methodology for gathering observational data on engaging messages, integrating the methodology proposed by Dale et al. (2022) with keyword-based filtering. In addition, we examined the engaging messages identified to observe their prevalence and evolution across educational levels and academic year. The method proved to be useful for analysing large amounts of text, as it allowed us to only have to examine 10% of the transcripts. Findings showed that teachers use all the types of messages, although in different proportions. *Identified* messages, both *gain-* and *loss-framed*, were the most common used messages in all grades, and across the academic year. The results also showed that, as the academic year progresses, the number of messages decreases. Thus, the present findings highlight the usefulness of the semi-automatic method for analysing teachers' discourse. This methodology not only streamlines the data analysis process but also provides comprehensive insights into the use and variation of engaging messages over time. Future research can build on this approach to further explore the dynamics of teacher-student interactions and develop targeted interventions to enhance educational outcomes.

**Acknowledgments**

None.

**Declaration of interest statement**

None.

**Tables**

**Table 1**
*Examples of theory-based engaging messages*

| Frame (Type of outcome highlighted) | Appeal (Type of motivational incentive) | Example |
|---|---|---|
| Gain (Emphasise the benefits of engaging in the task) | Extrinsic (External rewards such as privileges or tangible benefits) | *"If you pay attention in class, I'll let you spend the last few minutes of class on whatever you want."* |
| | Introjected (Internal pressures such as feelings of pride) | *"If you do your homework, you will feel satisfied."* |
| | Identified (Understanding and valuing the long-term benefits of the task) | *"If you work hard, you will be able to choose what to study in the future."* |
| | Intrinsic (Deriving inherent enjoyment and satisfaction from the task itself) | *"If you study, you will enjoy and have fun in this subject."* |
| Loss (Emphasise the disadvantages of not engaging in the task) | Extrinsic (External punishments such as losing privileges or facing penalties) | *"If you don't pay attention in class, you will be punished without recess."* |
| | Introjected (Internal pressures such as feelings of shame or disappointing others) | *"If you don't do your homework, you will disappoint me and your parents."* |
| | Identified (Understanding and valuing the long-term disadvantages of not engaging) | *"If you don't work hard, you'll have to make do with studying less sought-after degrees."* |
| | Intrinsic (Missing out on the inherent enjoyment and satisfaction of the task) | *"If you don't study, you will miss out on the beauty of this subject."* |

**Table 2**

*Engaging messages found in the audio recordings*

| Frame | Appeal | Example |
|---|---|---|
| Gain | Extrinsic | *"Well, let's do a quick review of ionic bonding to finish it off and then, if you're good, I'll play some music for you."* |
| Gain | Introjected | *"Guys do it right and try hard and you will be proud of yourselves after a job well done."* |
| Gain | Identified | *"The harder you try and the more you work now, the easier it will be in the future. Try to make a little effort, then it will pay off in the future."* |
| Gain | Intrinsic | *"Classes like this are enjoyable when you study and come here to ask questions."* |
| Loss | Extrinsic | *"If you don't want to pay attention, go to the corridor, come on, if you don't want to pay attention, go outside."* |
| Loss | Introjected | *"This really pisses me off. It's only the day before the exam that you ask questions."* |
| Loss | Identified | *"Can I do higher education? Of course. When? When my head focuses on what the person in front of me is explaining. If not, it will take me for ever."* |
| Loss | Intrinsic | *"Copying without understanding anything is a pain because you don't understand anything, and it is boring."* |

**Figures**

**Figure 1**. Three step procedure of the coding process

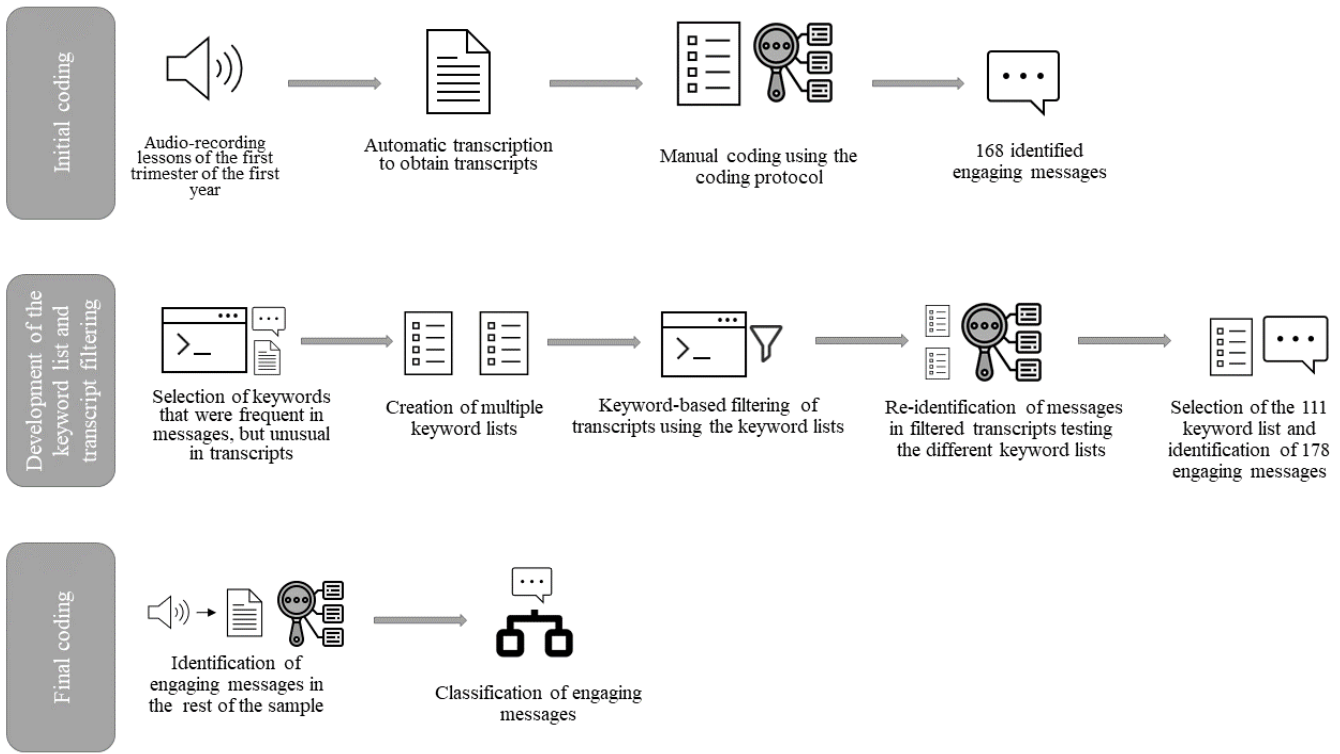

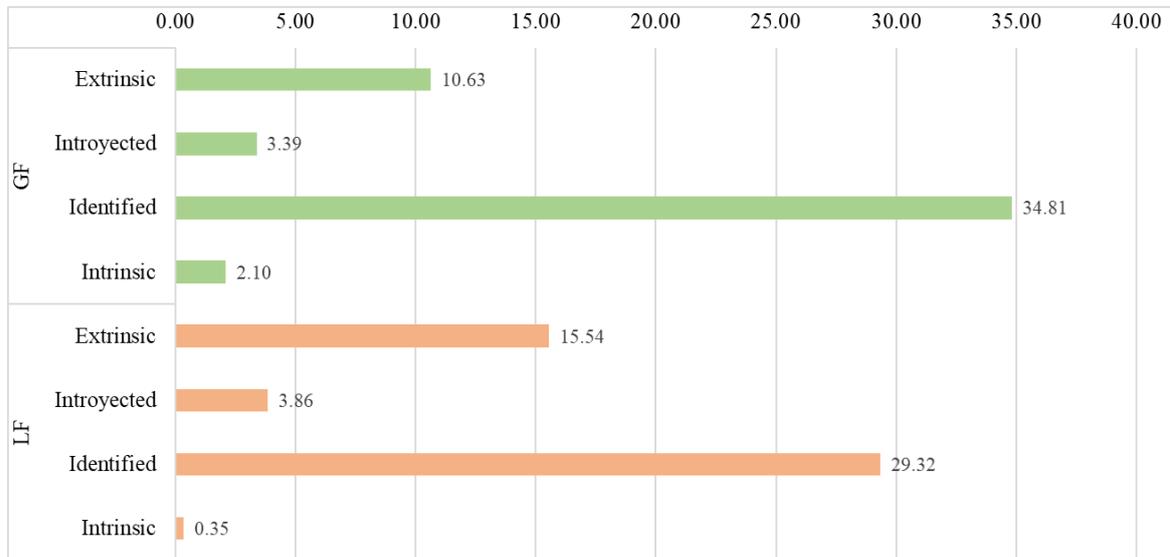

**Figure 2**. *Percentages of engaging messages found throughout the study*

*Note.* GF = Gain-framed (green); LF = Loss-framed (red).

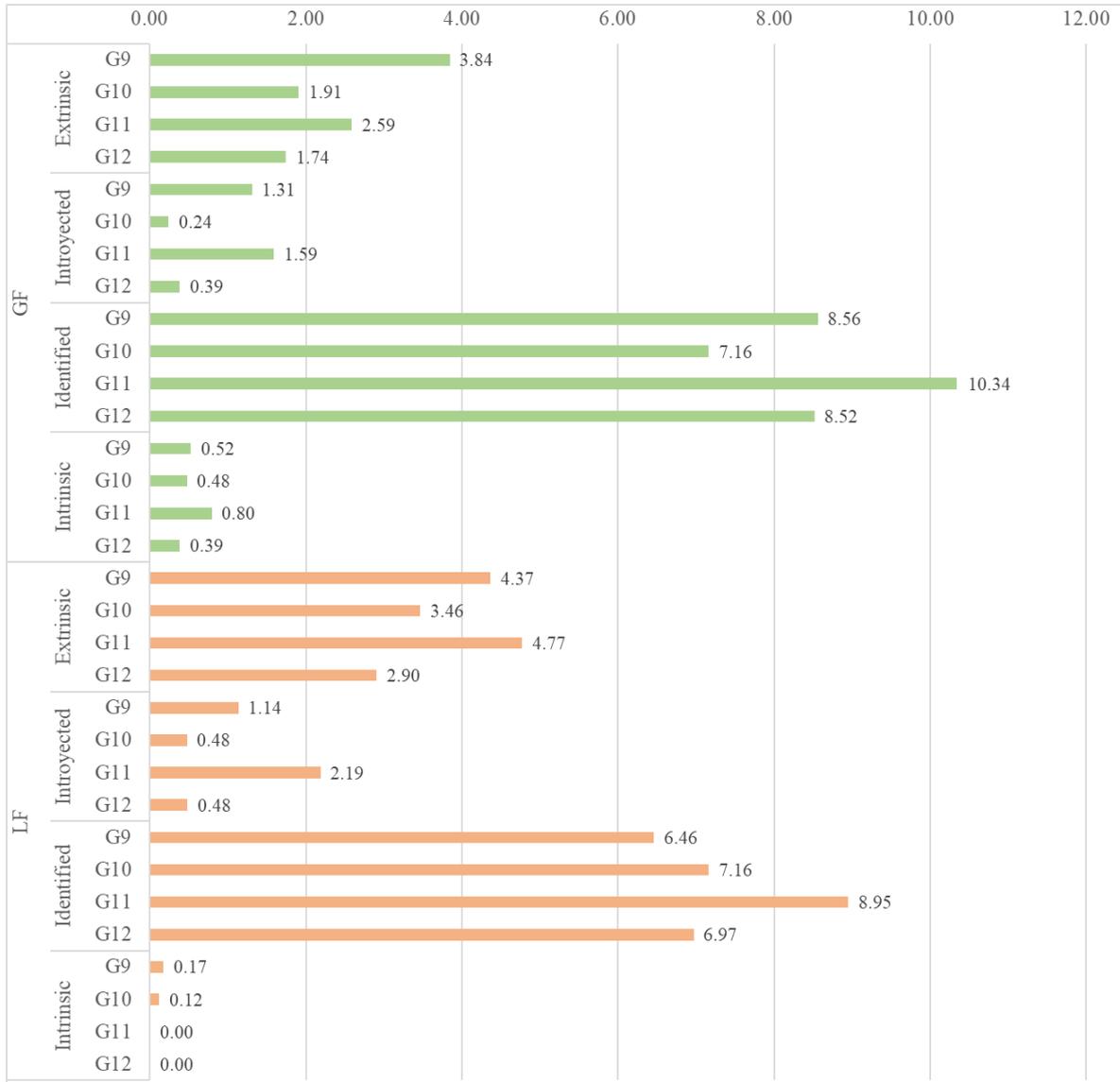

**Figure 3**. *Percentages of engaging messages found across educational levels*

*Note.* GF = Gain-framed (green); LF = Loss-framed (red); G9 = Grade 9; G10 = Grade 10; G11 = Grade 11; G12 = Grade 12.

**Figure 4**. *Percentages of engaging messages found throughout the trimesters*

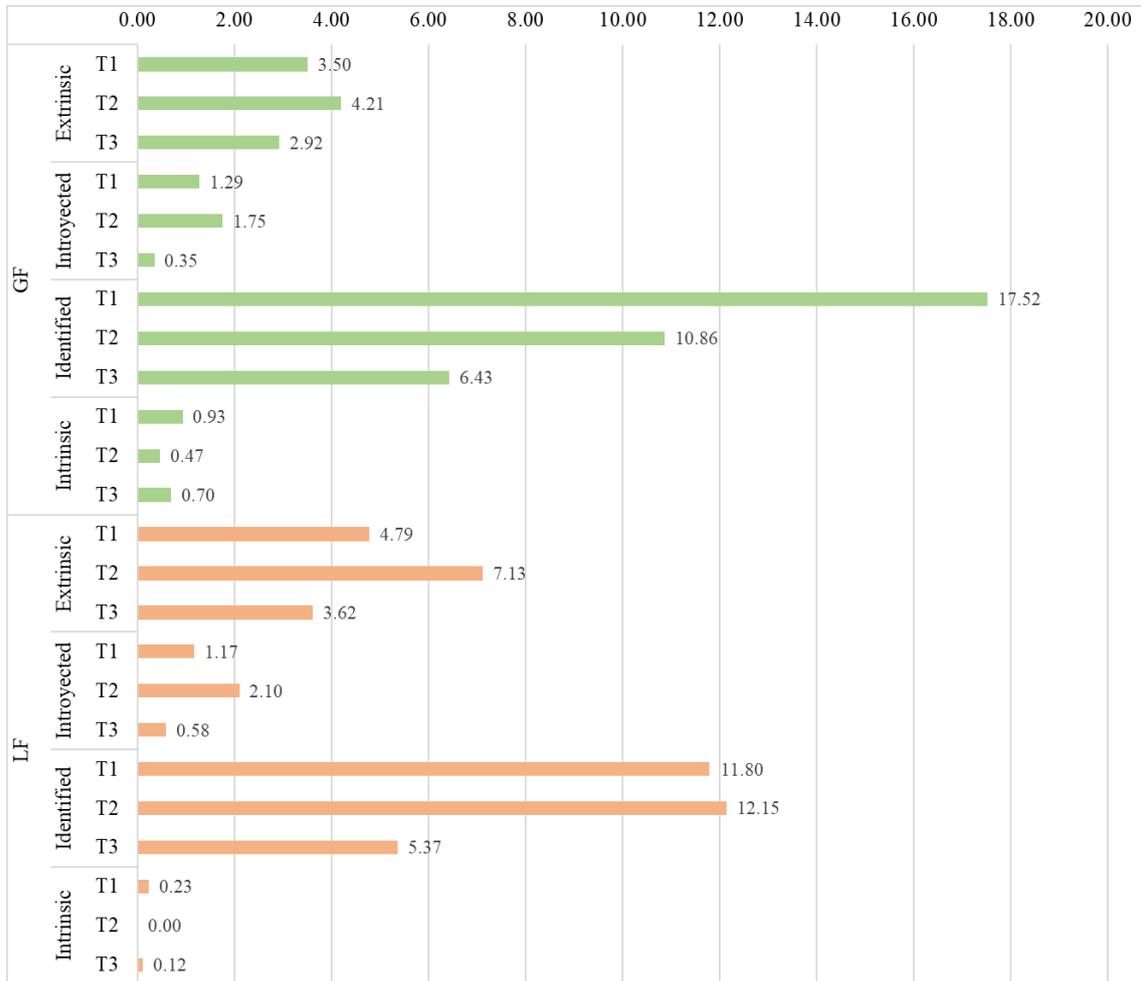

*Note.* GF = Gain-framed (green); LF = Loss-framed (red); T1 = First trimester; T2 = Second trimester; T3 = Third trimester.